\pdfoutput=1
\documentclass[11pt,dvipsnames]{article}

\usepackage{EMNLP2023}

\usepackage{times}
\usepackage{latexsym}

\usepackage[T1]{fontenc}
\usepackage[utf8]{inputenc}
\usepackage{microtype}
\usepackage{inconsolata}

\usepackage{amsmath}
\usepackage{graphicx}
\usepackage{xcolor}
\usepackage{enumerate}
\usepackage{booktabs}
\usepackage{multirow}
\usepackage{makecell}
\usepackage{xspace}
\usepackage{xurl}
\usepackage{hyperref}
\usepackage[noabbrev]{cleveref}

\title{Predictive Chemistry Augmented with Text Retrieval}

\author{Yujie Qian$^{\dagger\ddagger}$ ~ Zhening Li$^\ddagger$ ~ Zhengkai Tu$^\ddagger$ ~ Connor W. Coley$^{\ddagger\mathsection}$ ~  Regina Barzilay$^{\dagger\ddagger}$ \\
        $^\dagger$Computer Science and Artificial Intelligence Lab, MIT \\
        $^\ddagger$Department of Electrical Engineering and Computer Science, MIT \\
        $^\mathsection$Department of Chemical Engineering, MIT \\ 
        \texttt{\{yujieq,regina\}@csail.mit.edu} \quad \texttt{\{zli11010,ztu,ccoley\}@mit.edu}
        }
\usepackage{amsmath,amsfonts,bm}

\def\eqref#1{equation~\ref{#1}}
\def\1{\bm{1}}

\def\vt{{\bm{t}}}

\def\vx{{\bm{x}}}

\DeclareMathAlphabet{\mathsfit}{\encodingdefault}{\sfdefault}{m}{sl}
\SetMathAlphabet{\mathsfit}{bold}{\encodingdefault}{\sfdefault}{bx}{n}

\begin{document}
\maketitle

\newcommand{\textreact}{\text{TextReact}\xspace}
\newcommand{\textreacttf}{\text{TextReact}$_\textit{tf}$\xspace}
\newcommand{\textreacttb}{\text{TextReact}$_\textit{tb}$\xspace}

\newcommand{\yq}[1]{{\color{red}{[YQ: #1]}}}
\newcommand{\rev}[1]{{#1}}

\begin{abstract}
This paper focuses on using natural language descriptions to enhance predictive models in the chemistry field. Conventionally, chemoinformatics models are trained with extensive structured data manually extracted from the literature. In this paper, we introduce \textreact, a novel method that directly augments predictive chemistry with texts retrieved from the literature. \textreact retrieves text descriptions relevant for a given chemical reaction, and then aligns them with the molecular representation of the reaction. This alignment is enhanced via an auxiliary masked LM objective incorporated in the predictor training. We empirically validate the framework on two chemistry tasks: reaction condition recommendation and one-step retrosynthesis.  By leveraging text retrieval, \textreact significantly outperforms state-of-the-art chemoinformatics models trained solely on molecular data.
\end{abstract}

\section{Introduction}

In this paper, we propose a method for leveraging automatically retrieved textual knowledge to improve the predictive capacity of chemistry models.  These chemoinformatics models are utilized for a wide range of tasks, from analyzing properties of individual molecules to capturing their interactions in chemical reactions \cite{chemprop,segler2018planning,coley2018machine}. However,  standard approaches make these predictions operating solely on molecular encodings. Despite significant advances in neural molecular representations, the accuracy of these models still offers room for improvement. We hypothesize that their performance can be further enhanced using relevant information from the scientific literature.  

\begin{figure*}[t!]
    \centering
    \includegraphics[width=\linewidth]{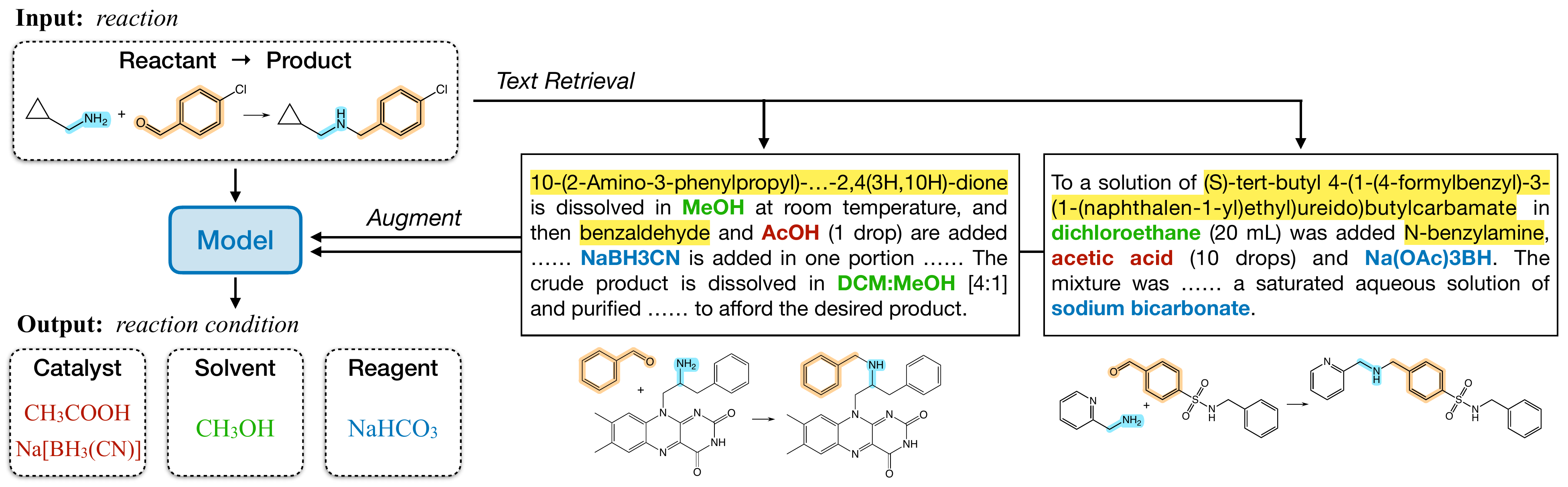}
    % \vspace{-0.1in}
    \caption{Predictive chemistry augmented with text retrieval. 
    For example, given the task of reaction condition recommendation, we retrieve texts relevant to the input reaction to provide additional evidence for the model's prediction. The two retrieved texts describe similar reactions to the input and share similar conditions. For visualization purposes, we mark the catalyst, solvent, and reagent in \textcolor{Maroon}{red}, \textcolor{Green}{green}, and \textcolor{RoyalBlue}{blue} in the retrieved texts. However, we do not assume that the text corpus contains such structured data.
    }
    \label{fig:example}
\end{figure*}

For instance, consider the task of finding a catalyst for the chemical reaction shown in Figure~\ref{fig:example}. This prediction proves to be challenging when considering the reaction components alone \cite{GaoFC20}.  At the same time, scientific literature provides several cues about potential catalysts (see highlighted excerpts).  While on their own these paragraphs might not provide a comprehensive answer, they can guide a molecular predictor. This intuition motivates our approach to aggregating textual sources with molecular representations when reasoning about chemistry tasks. Specifically, we aim to align the representations of the reaction and its corresponding text description, enabling the model to operate in the combined space. This design not only enables the model to retrieve readily available information from the literature but also enhances its generalization capacity for new chemical contexts.

% Databases of chemical reactions play an essential role in modeling molecular interactions and devising novel synthesis routes \cite{segler2018planning,coley2018machine,chemprop,predictive_chemistry}. Industry-leading databases such as Reaxys\footnote{\url{https://www.reaxys.com/}} and SciFinder\footnote{\url{https://scifinder.cas.org/}} predominantly rely on expert curation from published literature. One way to streamline this process is to develop information extraction models that automate the curation of reaction data \cite{LiuHTWF06,chemicaltagger,lowe2012extraction,cde}. However, this task poses considerable challenges due to the multifaceted and complex ways in which reactions are represented in publication, often incorporating both text and figures \cite{guo2021automated,RxnScribe,MolScribe}. 
% This paper introduces a different paradigm that retrieves texts from the literature on-demand for a specific chemistry task, thus augmenting the chemistry prediction model directly with natural language.

% \Cref{fig:example} illustrates our approach to reaction condition recommendation, where the task is to recommend catalysts, solvents, and reagents for a given chemical reaction. Traditionally, the input reaction is encoded with molecular fingerprints \cite{acscentsci.8b00357,acs.jcim.9b00313} or graph neural networks \cite{ryou2020graph}. We propose to retrieve texts that describe the same or similar reactions to the input, and provide them as additional evidence to support model prediction.  

%There are two main challenges for realizing %this approach. First, to effectively identify %texts relevant to a given reaction, we need to %align the representations of the reaction and %its corresponding text description. Second, %the prediction model must consolidate the %information from both the chemistry and %natural language space to provide a reliable %recommendation.

We propose \textreact, a novel predictive chemistry framework augmented with text retrieval. \textreact comprises two modules -- a SMILES-to-text retriever\footnote{SMILES, i.e.~Simplified Molecular-Input Line-Entry System, is a specification in the form of a line notation for describing chemical structures \cite{smiles}.} that maps an input reaction to corresponding text descriptions, and a text-augmented predictor that fuses the input reaction with the retrieved texts. The model learns the relation between a chemical reaction and text via an auxiliary masked LM objective incorporated in the predictor training. Furthermore, to improve generalization to unseen reaction classes, we simulate novel inputs by eliminating from the training data the closest textual descriptions for given reactions. 

In addition to condition recommendation, \textreact can be readily applied to other chemistry tasks. In our experiments, we also consider one-step retrosynthesis \cite{acscentsci.7b00355}, the task of predicting reactants used to synthesize a target molecule (see \Cref{fig:retro}).
 By leveraging relevant text, \textreact achieves substantial performance improvement compared to the state-of-the-art chemoinformatics models trained on reaction data alone. For instance, for condition recommendation, \textreact increases the top-1 prediction accuracy by 58.4\%, while the improvement in one-step retrosynthesis is 13.6--15.7\%. The improvement is consistent under both random and time-based splits of the datasets, validating the efficacy of our retrieval augmentation approach in generalizing to new task instances. 
\rev{Our code and data are publicly available at \url{https://github.com/thomas0809/textreact}.
}

\section{Related Work}

\paragraph{Multimodal Retrieval}
This field studies retrieval algorithms when the query and target are in different modalities, as exemplified by image-text retrieval \cite{weston2010large,Socher010,SocherKLMN14,KarpathyJL14,FaghriFKF18}. Previous research learns multimodal embeddings in image and text using techniques ranging from kernel methods to more expressive neural networks. This line also extends to other modalities such as video \cite{MiechASLSZ20} and audio \cite{aytar2017see}.

Our retrieval method closely relates to CLIP \cite{RadfordKHRGASAM21}, a multimodal model that leverages contrastive learning to align images with their corresponding natural language descriptions. \rev{Moving to the chemistry domain, \citet{text2mol} proposed Text2Mol to retrieve molecules using natural language queries. Our SMILES-to-text retriever operates in the opposite direction, i.e., we use the aligned embedding spaces to retrieve a text description given a reaction query. We then use the retrieved text to enhance chemistry prediction.
}

% \paragraph{Document Retrieval} 
% This field studies algorithms for retrieving relevant documents from a large corpus in response to an information need (i.e., a query). In the context of natural language processing, both the query and document are natural language text. Traditional methods represent them as bag-of-words vectors, such as TF-IDF \cite{Jones04} and  BM25 \cite{BM25}. More recently, researchers proposed neural networks to learn latent vector representations for retrieval tasks \cite{LeeCT19,KarpukhinOMLWEC20}. The query and document are independently encoded with separate encoders, which are trained by contrastive learning on labeled (query, document) pairs. Afterward, the encodings of all the documents in the corpus are pre-computed and indexed, allowing for efficient retrieval using maximum inner product search \cite{MussmannE16}. 

% Our work also focuses on retrieving relevant documents in response to a query, but the query is a chemistry input (e.g., a molecule or reaction) rather than natural language text. 

\paragraph{Retrieval-Augmented NLP}
There have been several studies on augmenting NLP models with retrieval. The idea is to retrieve relevant documents from a corpus and use them as additional context to the model. Earlier works on open-domain question answering adopted standalone retrievers to identify supporting paragraphs for the given questions \cite{ChenFWB17,LeeCT19,KarpukhinOMLWEC20}. \citet{REALM} jointly trained an end-to-end retriever with a language model, and \citet{RAG} extended the idea to general sequence-to-sequence generation.

\begin{figure}[t!]
    \centering
    \includegraphics[width=\linewidth]{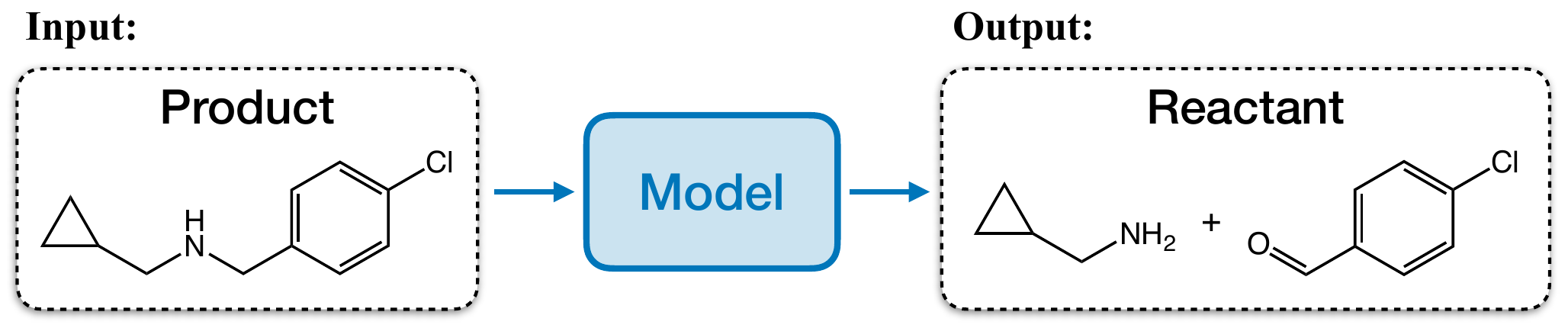}
    \caption{One-step retrosynthesis, another predictive chemistry task studied in this paper.}
    \label{fig:retro}
\end{figure}

Our work implements a similar idea of retrieval augmentation. However, we focus on tasks in the chemistry domain (see \Cref{fig:example,fig:retro}), with the goal of enhancing chemistry prediction models with natural language descriptions retrieved from the literature.

% \begin{figure}[t!]
%     \centering
%     \includegraphics[width=\linewidth]{figures/tasks.pdf}
%     \caption{Predictive chemistry tasks studied in this work. (a) Reaction condition recommendation. (b) One-step retrosynthesis.}
%     \label{fig:chemtask}
% \end{figure}

% \paragraph{Retrosynthesis}
% Retrosynthesis planning, a crucial task in the chemical and pharmaceutical industry, involves designing pathways for synthesizing target molecules from available starting chemicals. In recent years, machine learning methods have been applied in computer-aided synthesis planning (CASP) \cite{segler2018planning,doi:10.1126/science.aax1566,D0SC04184J}. These methods leverage neural networks to make predictions about chemical reactions, and use Monte Carlo tree search to explore and discover the retrosynthetic routes. In this work, we study two predictive chemistry tasks related to retrosynthesis, which are shown in \Cref{fig:chemtask}.

\begin{figure*}[t!]
    \centering
    \includegraphics[width=0.9\linewidth]{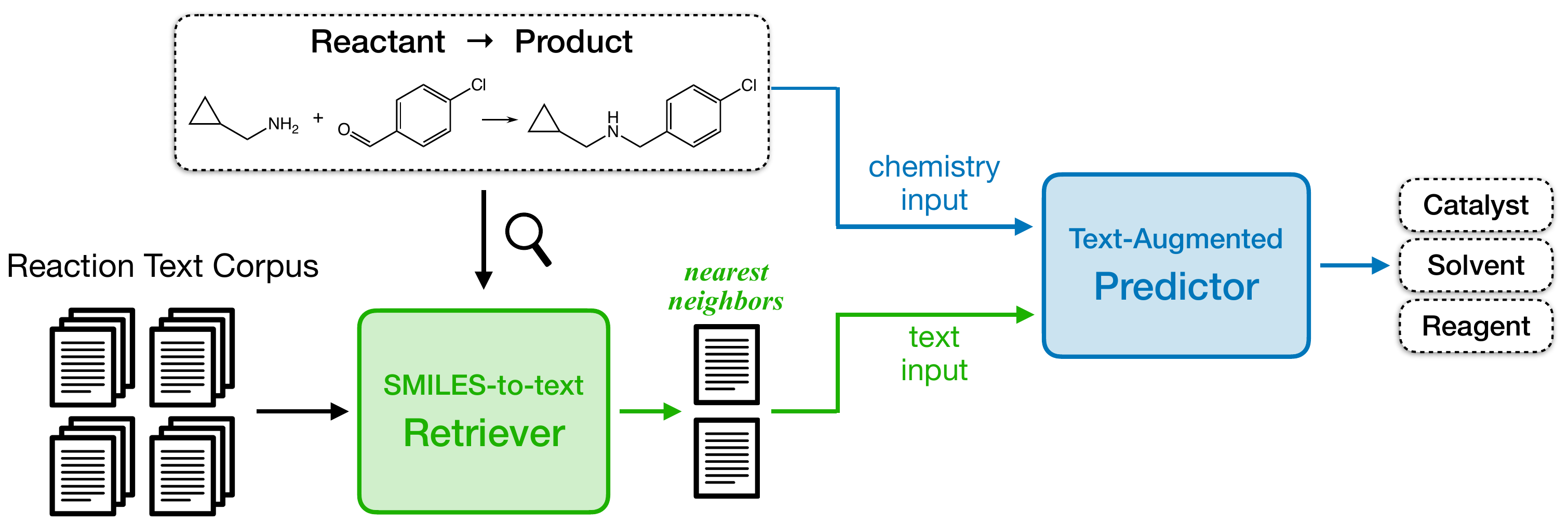}
    % \vspace{-0.1in}
    \caption{Overview of \textreact as applied to reaction condition recommendation. The retriever searches for texts relevant to the chemistry input, which are then used to augment the input of the predictor.  }
    \label{fig:textreact}
\end{figure*}

\paragraph{Natural Language Grounding}
% Our work relates to research on using natural language as an additional supervision signal for the model. Recent advancements in pretrained language models have demonstrated few-shot and zero-shot learning capability with the augmentation of natural language task descriptions \cite{gpt3,GaoFC20,WeiBZGYLDDL22}. 
Previous studies have demonstrated how natural language can be harnessed as a grounding mechanism to supervise tasks in various domains. For example, in computer vision, natural language descriptions have been used to improve fine-grained classification of bird images \cite{HeP17,LiangZY20} and to regularize the learning of visual representations \cite{AndreasKL18,MuLG20}. In reinforcement learning and robotics, natural language instructions are transformed into executable actions of the agent  \cite{BranavanCZB09,NarasimhanKB15,IchterBCFHHHIIJ22} and enable adaptability to new tasks \cite{HillLSCBMS20}. 

In the chemistry domain, our work complements ongoing efforts in the automatic extraction of structured reaction data from the chemistry literature~\cite{lowe2012extraction,cde,guo2021automated,RxnScribe,MolScribe,reactiondataextractor,HeNADTHAZFYACCB21,NguyenZYFDTHACB20,vaucher2020automated,ZhaiNATDCGV19}. While these works focus on creating structured databases used for training molecular models, we focus on directly leveraging unstructured natural language descriptions to further augment their predictions.

\section{Method}
\subsection{Problem Setup}

% Predictive chemistry relates to the development of machine learning models to describe chemical reactivity \cite{predictive_chemistry}. 
For concreteness, consider the task of reaction condition recommendation, 
where the input $X$ is a chemical reaction and the output $Y$ is a list of reaction conditions, including the catalyst, solvent, and reagent. We aim to train a machine learning model $\mathcal{F}$ to generate the prediction, i.e., $Y = \mathcal{F}(X)$. 
The model $\mathcal{F}$ is typically trained on a set of labeled training data $\mathcal{D}_\text{train}=\{(x_i, y_i),\ i=1,\dots,N\}$. 

In this paper, we incorporate two additional resources for the training of model $\mathcal{F}$:
\begin{enumerate}[(1)]
\item Each example in the training set is paired with a text paragraph, i.e. $(x_i, y_i, t_i)$, where $t_i$ describes the corresponding chemical reaction. Many reaction databases are curated from the chemistry literature and provide text references for their reaction data.\footnote{Examples of such text references are available in the appendix.}
\item An unlabeled text corpus of chemical reactions $\mathcal{T}=\{t_j,\ j=1,\dots,M\}$, where each $t_j$ is a paragraph describing a chemical reaction. They can be easily obtained from the text of journal articles and patents using a model or heuristics to determine that they contain a reaction description.
\end{enumerate}

More generally, other predictive chemistry tasks can use the same problem setup while changing the definitions of $X$ and $Y$ accordingly. For instance, in one-step retrosynthesis, $X$ is a product molecule and $Y$ is a list of potential reactants.

% These resources are in principle readily available, as many reaction databases are curated from the chemistry literature and provide text references for the reaction data. Additionally, the unlabeled text corpus can be obtained from chemistry journal articles and patents.

\subsection{TextReact Framework}

We propose a novel framework \textreact to augment chemistry prediction with text retrieval. As illustrated in \Cref{fig:textreact}, the \textreact framework comprises two major components: a SMILES-to-text retriever (\Cref{sec:retriever}) and a text-augmented predictor (\Cref{sec:predictor}). The retriever searches the unlabeled corpus for texts relevant to the particular chemistry input. The predictor leverages both the chemistry input and the retrieved texts to generate the prediction.

\subsubsection{SMILES-To-Text Retriever}
\label{sec:retriever}

The goal of the retriever is to locate relevant texts from an unlabeled corpus based on a given chemistry input. 
To accomplish this task, we devise a SMILES-to-text retriever, leveraging the widely used dual encoder architecture employed in document retrieval \cite{LeeCT19,KarpukhinOMLWEC20} and image-text retrieval \cite{RadfordKHRGASAM21}. The model consists of two parts: a \textit{chemistry encoder}  for processing the input molecule or reaction represented as a SMILES string, and a \textit{text encoder} for encoding the text descriptions. For the chemistry encoder, we use a Transformer to encode the SMILES string into a latent vector. As for the text encoder, we employ a Transformer pre-trained on scientific text \cite{scibert} to encode each paragraph into a latent vector. To enable efficient retrieval using nearest neighbor search, we train the model to align the two latent spaces.
% Simplified Molecular-Input Line-Entry System (SMILES) string\footnote{SMILES is a specification in the form of a line notation for describing chemical structure \cite{smiles}. The SMILES string for a molecule is derived through a depth-first traversal of the corresponding graph. The reaction SMILES is defined as ``reactant>reagent>product''. }, 

\begin{figure}[t!]
    \centering
    \includegraphics[width=\linewidth]{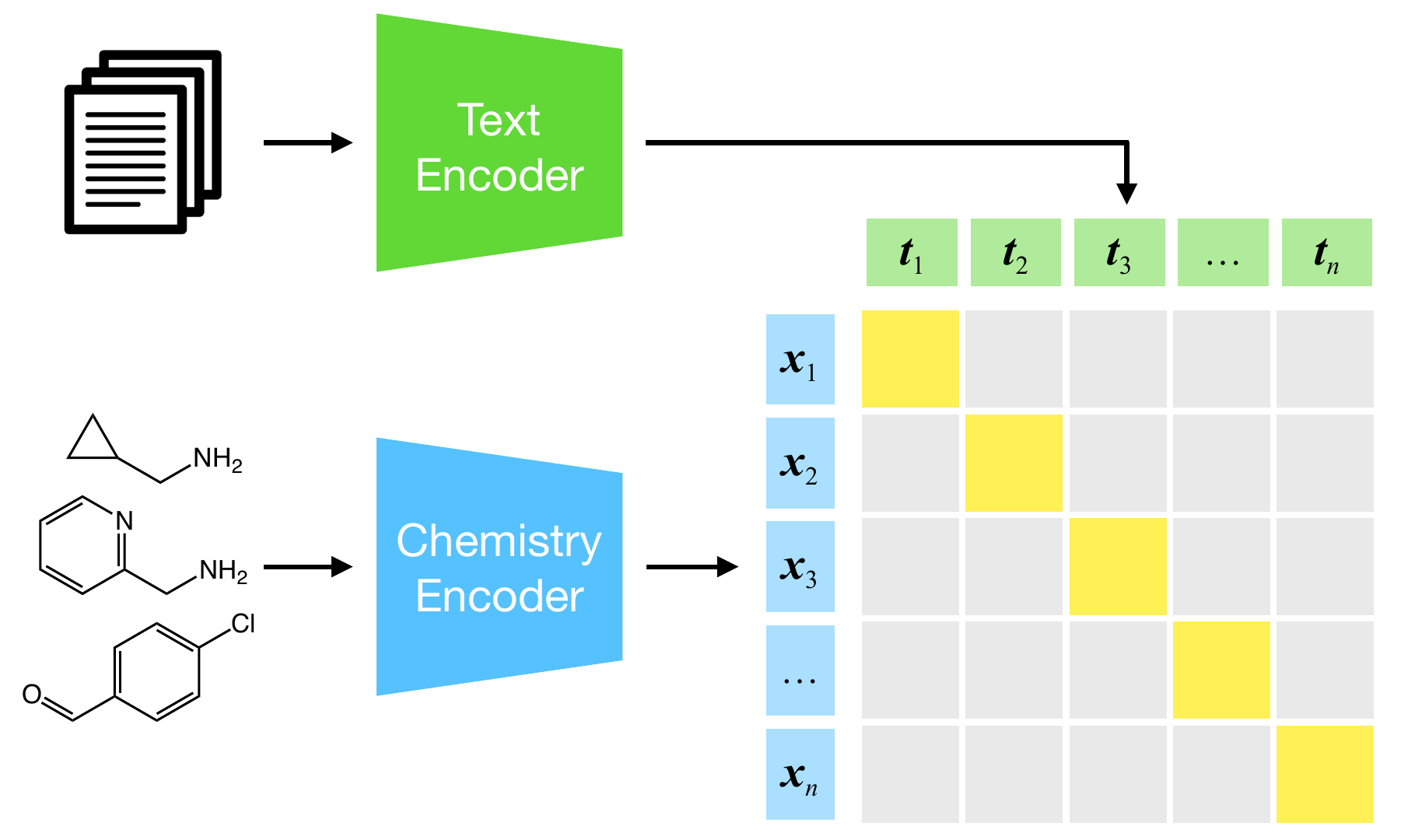}
    \caption{SMILES-to-text retriever trained with contrastive learning.}
    \label{fig:retrieve}
\end{figure}

The retriever is trained with contrastive learning, as illustrated in \Cref{fig:retrieve}. Given a batch of SMILES strings and their corresponding paragraphs $\{(x_i, t_i), i=1,\dots,n\}$, we first compute their encodings,
\begin{equation}
    \vx_i = \mathrm{ChemEnc}(x_i), \quad
    \vt_i = \mathrm{TextEnc}(t_i).
\end{equation}
The similarity between a SMILES and a text paragraph is defined by the dot product of their encodings, 
\begin{equation}
    S_{i,j} = \vx_i ^\top \vt_j.
\end{equation}
For each SMILES string $x_i$, its paired paragraph $t_i$ is a positive example, and the other paragraphs within the same batch are considered negative examples. We further randomly sample $n$ paragraphs from the unlabeled corpus as additional negative examples for each batch, denoted as $t_{n+1}, \dots, t_{2n}$. The training objective is to maximize the log-likelihood of matching the SMILES strings with positive text paragraphs, i.e.,
\begin{equation}
    L_\text{ret}=- \sum_{i=1\dots n} \log \frac{\exp(S_{i,i})}{\sum_{j=1\dots 2n} \exp(S_{i,j})}.
\end{equation}

After training, we pre-compute all text encodings and compile them into an index to support efficient retrieval. When retrieving relevant texts for a given SMILES string, we compute its encoding with the chemistry encoder and then run a maximum inner product search \cite{MussmannE16,johnson2019billion} to find its nearest neighbors in the index.

In this paper, we train the retriever separately from the predictor with a standalone objective. While joint optimization of the retriever and the predictor is possible \cite{REALM,RAG}, it requires significantly more computation as it involves iterative index rebuilding and retrieval during training. We choose a standalone retriever that has exhibited strong performance in identifying relevant texts in our experiments (see \Cref{tab:retriever}).

\subsubsection{Text-Augmented Predictor}
\label{sec:predictor}
% We propose a novel model for predictive chemistry that incorporates text augmentation, using both the chemistry input and retrieved texts to generate predictions. 
The goal of this component is to merge chemistry and natural language to produce accurate predictions.
Our text-augmented predictor is designed as an encoder-decoder model, with the encoder handling both the chemistry input and retrieved texts, and the decoder generating the corresponding chemistry output. 

The encoder's input consists of the concatenation of the chemistry input (i.e., the SMILES string) and the retrieved texts. To structure the input, we prepend a \texttt{[CLS]} token at the beginning and use \texttt{[SEP]} tokens to separate the SMILES and the texts. This yields the following input format:
\begin{equation*}
\resizebox{\hsize}{!}{$
    \texttt{[CLS]}\ \texttt{SMILES}\ \texttt{[SEP]}\ (0)\ \texttt{TEXT}_0 \ (1)\ \texttt{TEXT}_1 \cdots \texttt{[SEP]}
$}
\end{equation*}
\texttt{TEXT}$_0$, \texttt{TEXT}$_1$, $\cdots$ are the nearest neighbors retrieved by the SMILES-to-text retriever, which serve as additional input to augment chemistry prediction. The number of the appended nearest neighbors $k$ is a hyperparameter.

The decoder architecture is tailored to the specific predictive chemistry task. For reaction condition recommendation, we adopt the approach established in prior research~\cite{acscentsci.8b00357}, which generates the reaction conditions in a specific order: catalyst, solvent~1, solvent~2, reagent~1, and reagent~2. For one-step retrosynthesis, the decoder \rev{either directly generates the SMILES strings of the reactants (i.e., a \textit{template-free} approach), or predicts the reaction template first and uses cheminformatics software to derive the reactants (i.e., a \textit{template-based} approach).}\footnote{\rev{Depending on whether to use an explicit set of reaction templates, existing one-step retrosynthesis models can be categorized into \textit{template-based} and \textit{template-free} approaches. See \Cref{sec:retro} and \Cref{appendix:retro} for further discussion.}} 
% More details about the decoder are given in \Cref{sec:implementation}.

The predictor is trained via supervised learning to maximize $p(y_i\mid x_i, T_i)$,
where $T_i$ is the set of retrieved texts.
To encourage the predictor to leverage the information of both chemistry and text, we add an auxiliary masked language model (MLM) loss. Specifically, we randomly mask out portions of the input, either the SMILES string or the text, and add a prediction head on top of the last layer of the encoder to predict the masked tokens. Inspired by previous research on language model pre-training \cite{BERT,BART,SpanBERT,GLM}, the masking is performed by repeatedly sampling spans of length drawn from a Poisson distribution with $\lambda=3$, until the total number of masked tokens reaches 15\% of the original input. With the MLM loss, the model learns to infer the missing part of a SMILES string from its nearest neighbor texts, and vice versa, thus encouraging it to learn the correspondence between chemistry and text. The masking also encourages the predictor not to rely on a single source of input, similar to dropout, thus improving its robustness. Finally, the training objective of the predictor is 
\begin{equation}
    L_\text{pred} = - \sum_i \log p(y_i\mid x_i, T_i) + \lambda_1 L_\text{MLM}
\end{equation}
where $\lambda_1$ is a hyperparameter controlling the weight of the MLM loss $L_\text{MLM}$.

To improve the generalizability of the predictor, we employ a dynamic sampling training strategy
that addresses the following important distinction between the training and testing of the predictor. During training, we have access to the \textit{gold text} paired with each chemistry input, providing a description of the corresponding reaction. However, in testing, it is not guaranteed to retrieve the gold text. Moreover, for novel chemistry inputs lacking a corresponding text description in the existing corpus, the predictor can only use information from the nearest neighbors, which may describe similar but not identical reactions. To address this gap, we employ a random sampling policy during training to simulate novel chemistry inputs that are not present in the corpus. With probability $\alpha$, the chemistry input is augmented with $k$ random neighbors from its top-$K$ nearest neighbors, where $K>k$ and the top-$K$ neighbors are expected to cover reactions similar to the chemistry input. In the remaining cases, the model is given
the gold text along with the top-$(k-1)$ neighbors (excluding the gold text) returned by the retriever. During inference, the examples are always augmented with the top-$k$ nearest neighbors.

\section{Experimental Setup}
\paragraph{Text Corpus}
We construct a text corpus from USPTO patent data processed by \citet{lowe2012extraction}. Each paragraph in the corpus provides a description of the synthesis procedure of a chemical reaction. The corpus consists of 2.9 million paragraphs, with an average length of 190 tokens.\footnote{We use the tokenizer of SciBERT: \url{https://huggingface.co/allenai/scibert_scivocab_uncased}.} While the original dataset includes structured reaction data extracted from each paragraph, including reactants, products, and reaction conditions, our model does not rely on this extracted data but instead learns directly from the unlabeled text.

\paragraph{Implementation} 
\label{sec:implementation}
The SMILES-to-text retriever is implemented with Tevatron \cite{tevatron}.\footnote{As the original Tevatron toolkit does not support different encoders and tokenizers for queries and passages, we make modifications to accommodate these requirements.} The chemistry encoder is initialized with ChemBERTa \cite{chemberta} (pre-trained on the SMILES strings from a molecule database) and the text encoder is initialized with SciBERT \cite{scibert} (pre-trained on scientific text from the literature). We finetune a separate retriever on the training set of each chemistry task. The input to the chemistry encoder is slightly different in each task. For reaction condition recommendation, the input is the reactants and product of a reaction, while for one-step retrosynthesis, the input is only the product. In both cases, the chemistry input is represented as a SMILES string. 

\rev{The text-augmented predictor employs a pre-trained SciBERT \cite{scibert} as the encoder.}
We concatenate the input SMILES string with three neighboring text paragraphs ($k=3$) by default, using \verb+[CLS]+ and \verb+[SEP]+ tokens as described in \Cref{sec:predictor}. During training, we set the random sampling ratio to $\alpha=0.8$ for reaction condition recommendation and $\alpha=0.2$ for one-step retrosynthesis, and the cut-off $K$ is set to 10. We analyze the effect of the hyperparameters in \Cref{sec:analysis}. 
% For reaction condition recommendation, we follow previous work \cite{acscentsci.8b00357} and use the decoder to sequentially predict the catalyst, solvent 1, solvent 2, reagent 1, and reagent 2, where each prediction is treated as a classification problem. For one-step retrosynthesis, the decoder generates the SMILES string of the reactants given the target product. The SMILES string is tokenized using the tokenizer developed by \citet{schwaller2021mapping}.  
During inference, we use beam search to derive the top predictions. 
More implementation details can be found in \Cref{appendix:implement}.

\paragraph{Evaluation} 
For each chemistry task, we evaluate \textreact under two setups. The first setup is the \textit{random split}, where the dataset is randomly divided into training/validation/testing. This is a commonly used setup in previous chemistry research \cite{acscentsci.8b00357,acscentsci.7b00355}. The second and more challenging setup is the \textit{time split}, where the dataset is split based on the patent year. We train the model with historical data from older patents and test its performance on the data from newer patents. Due to the substantial differences between reactions in new patents and previous ones, it becomes more challenging for a model to generalize effectively under a time split. During testing, we compare retrieving from the full corpus (including newer patents) and retrieving only from the years used for training. \Cref{tab:stats} shows the statistics of the datasets for the two tasks, which are elaborated in the next section.

\begin{table}[t!]
    \centering
    \setlength{\tabcolsep}{4pt}
    \begin{tabular}{crrrr}
        \toprule
         & \makecell{RCR\\(RS)} & \makecell{RCR\\(TS)} & \makecell{RetroSyn\\(RS)} & \makecell{RetroSyn \\(TS)} \\\midrule
        Train & 546,728 & 565,575 & 40,008 & 38,631 \\
        Valid & 68,341  & 63,015  & 5,001  & 5,624 \\
        Test  & 68,341  & 54,820  & 5,007  & 5,761 \\
        % Corpus & 2,947,185 & 2,505,440 & 2,947,185 \\
        \bottomrule
    \end{tabular}
    \caption{Statistics of datasets for reaction condition recommendation (RCR) and one-step retrosynthesis (\text{RetroSyn}). RS: random split; TS: time split. }
    \label{tab:stats}
\end{table}

\begin{table}[t!]
    \centering
    \begin{tabular}{lccccccc}
        \toprule
         & R@1 & R@3 & R@10  \\\midrule
         RCR (RS) & 70.9 & 91.6 & 97.0 \\
         RCR (TS) & 75.1 & 91.8 & 96.2 \\
         RetroSyn (RS) & 60.0 & 82.5 & 92.5 \\
         RetroSyn (TS) & 58.0 & 81.4 & 91.5 \\
         \bottomrule
    \end{tabular}
    \caption{Performance of our SMILES-to-text retriever trained on each dataset. We report the Recall@\{1,3,10\} when retrieving from the full corpus. Scores are in \%.}
    \label{tab:retriever}
\end{table}

\begin{table*}[t!]
    \centering
    \setlength{\tabcolsep}{4.5pt}
    \begin{tabular}{lccccccccc}
        \toprule
            & \multicolumn{4}{c}{RCR (RS)} && \multicolumn{4}{c}{RCR (TS)} \\\cmidrule{2-5} \cmidrule{7-10}
            & Top-1 & Top-3 & Top-10 & Top-15 && Top-1 & Top-3 & Top-10 & Top-15 \\\midrule
        rxnfp LSTM \cite{acscentsci.8b00357} 
            & 20.5 & 30.7 & 41.7 & 45.3 && 15.2 & 26.2 & 40.7 & 45.4 \\
        rxnfp retrieval 
            & 27.2 & 37.5 & 47.9 & 51.1 && 7.8  & 15.2 & 27.3 & 31.5 \\
        Transformer 
            & 30.0 & 43.8 & 56.7 & 60.5 && 18.7 & 31.8 & 47.6 & 52.7 \\
        ChemBERTa
            & 30.3 & 44.7 & 58.0 & 62.0 && 18.7 & 31.9 & 47.6 & 52.8 \\ 
            %\midrule
        TextReact
            % & \\
        % $\ \cdot$ full corpus 
            & \textbf{88.4} & \textbf{93.9} & \textbf{96.0} & \textbf{96.5} && \textbf{83.9} & \textbf{90.9} & \textbf{93.9} & \textbf{94.6} \\
        % $\ \cdot$ gold-removed
        %     & 47.2 & 59.9 & 69.0 & 72.5 && 36.3 & 50.4 & 63.8 & 67.9 \\
        % $\ \cdot$ TS corpus
        %     & ---  & ---  & ---  & ---  && 21.1 & 35.2 & 51.0 & 56.1 \\
        \bottomrule
    \end{tabular}
    \caption{Evaluation results for reaction condition recommendation (RCR). RS: random split; TS: time split. Scores are accuracy in \%. }
    \label{tab:rcr}
\end{table*}

\begin{table*}[t!]
    \centering
    \setlength{\tabcolsep}{4pt}
    \begin{tabular}{lccccccccc}
        \toprule
            & \multicolumn{4}{c}{RetroSyn (RS)} && \multicolumn{4}{c}{RetroSyn (TS)} \\\cmidrule{2-5} \cmidrule{7-10}
            & Top-1 & Top-3 & Top-5 & Top-10 && Top-1 & Top-3 & Top-5 & Top-10 \\\midrule
        \small\textit{Template-free models} \\
        ~ G2G \cite{ShiXG0T20}
            & 48.9 & 67.6 & 72.5 & 75.5 && --- & --- & --- & --- \\
        ~ Transformer \cite{C9SC03666K}
            & 43.1 & 64.6 & 71.8 & 78.7 && --- & --- & --- & --- \\
        ~ Dual-TF \cite{sun2020energy}
            & 53.6 & 70.7 & 74.6 & 77.0 && --- & --- & --- & --- \\
        ~ Transformer$_\mathit{tf}$ 
            & 45.9 & 63.2 & 68.5 & 75.2 && 31.4 & 46.6 & 51.6 & 57.2 \\
            % \midrule
        ~ TextReact$_\mathit{tf}$
        %     & \\
        % $\ \cdot$ full corpus 
            & \textbf{59.5} & \textbf{72.6} & \textbf{75.9} & \textbf{80.0} && \textbf{51.0} & \textbf{64.1} & \textbf{68.1} & \textbf{72.9} \\
        % $\ \cdot$ gold-removed
        %     & 47.7 & 62.8 & 67.7 & 72.6 && 40.1 & 54.1 & 58.8 & 63.9 \\
        % $\ \cdot$ TS corpus
        %     & ---  & ---  & ---  & ---  && 31.3 & 44.8 & 50.0 & 55.6 \\
            \midrule
        \small\textit{Template-based models} \\
        ~ LocalRetro \cite{localretro}
            & 53.4 & 77.5 & 85.9 & 92.4 && --- & --- & --- & --- \\
        ~ O-GNN \cite{ZhuWWX0MWQZLL23}
            & 54.1 & 77.7 & 86.0 & \textbf{92.5} && --- & --- & --- & --- \\
        ~ Transformer$_\mathit{tb}$
            & 52.5 & 72.8 & 79.7 & 86.2 && 43.6 & 65.6 & 73.2 & 82.1 \\
        ~ TextReact$_\mathit{tb}$ 
            & \textbf{68.2} & \textbf{83.7} & \textbf{88.1} & \textbf{92.5} && \textbf{68.7} & \textbf{84.5} & \textbf{88.8} & \textbf{92.8} \\
        \bottomrule
    \end{tabular}
    \caption{Evaluation results for one-step retrosynthesis. RS: random split; TS: time split. Scores are accuracy in \%.}
    \label{tab:retro}
\end{table*}

\section{Experiments}

\subsection{Reaction Condition Recommendation}
\label{sec:rcr}

\paragraph{Data}
We follow the setup of previous research \cite{acscentsci.8b00357} to construct reaction condition datasets from the USPTO data \cite{lowe2012extraction}. The reactions with at most one catalyst, two solvents, and two reagents are kept, and reactions with conditions that occurred fewer than 100 times are excluded.\footnote{\citeauthor{acscentsci.8b00357} constructed their datasets from Reaxys, which are not publicly available. We use the preprocessing script of Parrot (\url{https://github.com/wangxr0526/Parrot}) to process the public USPTO data.} 
Each reaction is associated with a text paragraph (gold) from the patent. However, we only utilize the gold text during the training of the retriever and the predictor, and do not use it for validation or testing.

% \begin{itemize}
%     \item RCR (RS), where the reactions are \textit{randomly split} into training/validation/testing with a ratio of 80\%/10\%/10\%.
%     \item RCR (TS), where we perform a \textit{time split} of the dataset. The reactions collected from patents before 2015 are categorized as the training set, reactions from 2015 as validation, and reactions from 2016 as testing.
% \end{itemize}

% \Cref{tab:stats} shows the statistics of the datasets. 

We create two splits of the dataset:  RCR (RS) for \textit{random split}, and RCR (TS) for \textit{time split}. More details can be found in \Cref{appendix:dataset}.

\paragraph{Baselines}
We implement four baselines, none of which use additional text input:
\begin{itemize}
    \item Reaction fingerprint (rxnfp) LSTM, a reproduction of the method proposed by \citet{acscentsci.8b00357}. The reaction fingerprint is calculated as the difference between the product and reactant fingerprints, which is further encoded by a two-layer neural network, and an LSTM decodes the reaction conditions.
    \item Reaction fingerprint (rxnfp) retrieval, which uses the conditions of the most similar reactions in the training set as the prediction. Similar reactions are determined based on the $L_2$ distance of reaction fingerprints. This baseline examines the performance of a pure retrieval method on this task.
    \item Transformer, the most important baseline we are comparing with, which uses the same architecture as our predictor. This baseline represents the state-of-the-art model that only takes chemistry input.
    \item ChemBERTa \cite{chemberta}. This baseline is the same as the Transformer baseline except that the encoder is pretrained on external SMILES data. The purpose of this baseline is to demonstrate the impact of such pretraining.
\end{itemize}

\paragraph{Results}
\Cref{tab:rcr} shows our experimental results for reaction condition recommendation. 
\textreact substantially outperforms standard chemistry models (Transformer and ChemBERTa), which are trained on reaction data without text. This significant improvement can be attributed to the strong performance of our retriever (shown in \Cref{tab:retriever}). When the retriever successfully identifies the gold text, \textreact effectively utilizes this information to make predictions. Otherwise, \textreact can also benefit from retrieving texts of similar reactions, as we will demonstrate in \Cref{sec:analysis}.\footnote{Another noteworthy observation is that the rxnfp retrieval baseline performs comparably to Transformer, suggesting that similar reactions often share similar reaction conditions, validating the efficacy of retrieval-based methods.}

\textreact generalizes to a more challenging time split. 
As reactions from the same patent are often similar, it may be easy for the model to infer the reaction conditions when similar reactions are present in the training set. Unsurprisingly, all baselines perform worse under the time split, highlighting the inherent difficulty of achieving such generalization. In \textreact, despite being trained only on historical data, the retriever retains high accuracy in retrieving the corresponding paragraph for reactions in the testing set (see \Cref{tab:retriever}). Thus, \textreact retains a strong overall performance by leveraging the retrieved text. 

% In the hardest setting, where the retriever could only retrieve from the text corpus for the training set and cannot access the gold texts of reactions from the same patent as the testing examples, \textreact still outperforms the baselines by a significant margin (see the ``TS corpus'' row in \Cref{tab:rcr}).

% In a harder setting where the gold texts for the testing examples are excluded from the corpus (referred to as ``gold-removed'' in \Cref{tab:rcr}), \textreact still scores significantly higher than all the baselines. This finding highlights the advantage of leveraging text retrieval for novel reactions that lack descriptions in the corpus.  

% Under the random split, our Transformer baseline achieves a top-1 accuracy of 30\%, outperforming previous fingerprint-based method by 10\%. The pre-trained ChemBERTa model shows slightly better performance. Interestingly, the rxnfp retrieval baseline achieves a top-1 accuracy of 27.2\%, which is close to Transformer's performance. This result suggests that similar reactions often share similar reaction conditions, implying the potential of retrieval-based methods. With the augmented text, \textreact demonstrates a remarkable improvement in top-1 accuracy, achieving 88.4\%. 

% To test the generalizability of our model, we further evaluate it on a more challenging time split. 

\subsection{One-Step Retrosynthesis}
\label{sec:retro}

\paragraph{Data}
We use the popular USPTO-50K dataset \cite{acscentsci.7b00355} for our one-step retrosynthesis experiment. %Similar to \Cref{sec:rcr}, 
We also create two data splits: \text{RetroSyn (RS)}, the original \textit{random split} of the dataset, and RetroSyn (TS), the \textit{time split}. 

Since the dataset was constructed from the same USPTO data as our text corpus, we match the examples in the dataset with text paragraphs in the corpus. However, due to differences in preprocessing, not all examples could be matched.\footnote{ Out of the 40,008 examples in the training set, 31,391 are matched.} We use only the matched examples to train the retriever,  whereas the predictor is trained with the full dataset. 

\paragraph{Template-free \& Template-based Models}
\rev{
We implement \textreact in two settings: (1) \textreacttf is a template-free model that uses a Transformer decoder to generate the SMILES strings of reactants. (2) \textreacttb is a template-based model that follows the formulation of \text{LocalRetro} \cite{localretro}. Specifically, we adopt the set of reaction templates extracted from the training data, and predict which template is applicable to a product molecule and which atom or bond is the reaction center. For each atom, we represent it using the corresponding hidden state from the last layer of the Transformer encoder and predict a probability distribution over the reaction templates using a linear head. For each bond, we concatenate the representations of its two atoms and employ another linear head to predict the template.
}

\begin{figure*}[t]
    \centering
    \includegraphics[width=\linewidth]{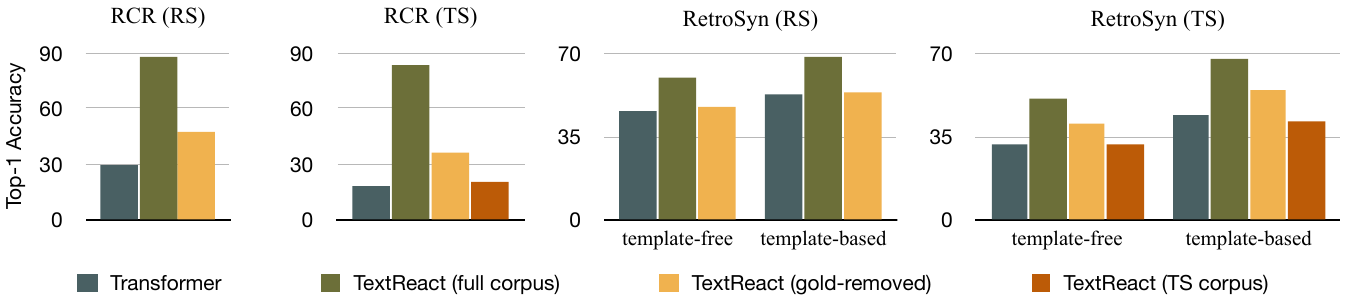}
    \caption{Comparison of \textreact's performance under three settings: (1) Retrieving from the full corpus, (2) Retrieving from the corpus with gold texts of testing examples removed, and (3) Retrieving only from the historical corpus under a time split. }
    \label{fig:corpuseval}
\end{figure*}

\paragraph{Baselines}
\rev{
We implement baselines Transformer$_\textit{tf}$ and Transformer$_\textit{tb}$ for template-free and template-based settings, respectively. They closely follow the output format of \textreact but do not incorporate retrieved text as additional input. We also report the published results of previous research \cite{ShiXG0T20,C9SC03666K,sun2020energy,localretro,ZhuWWX0MWQZLL23}. 
All baselines are trained on the same data as \textreact, but without the text input. 
}

\paragraph{Results}
\Cref{tab:retro} shows the results of one-step retrosynthesis. Similar to the RCR task, \textreact has demonstrated strong performance by leveraging text retrieval. \rev{On the RetroSyn (RS) dataset, our baseline Transformer models perform comparably to previous models under both template-free and template-based settings. Upon integrating text augmentation, \textreacttf and \textreacttb advance the top-1 accuracy by 13.6\% and 15.7\%, respectively, affirming the advantage of retrieval augmentation for this task. }
Even under a time-split scenario, \textreact maintains a high accuracy, underscoring its proficiency in both retrieval and final prediction stages.

\subsection{Analysis}
\label{sec:analysis}

First, we demonstrate in \Cref{fig:corpuseval} that \textreact exhibits generalization capabilities to novel reactions not present within the text corpus. While we have illustrated the accurate retrieval of text descriptions from the corpus and their effective utilization for predictions, it is important to note that a gold text for the target reaction may not always be available within the corpus. To further assess the model's performance in a more challenging scenario, we remove the gold texts of all testing examples from the corpus (referred to as ``gold-removed'' in \Cref{fig:corpuseval}). In both RCR and RetroSyn, \textreact continues to outperform the Transformer baseline significantly. The improvement is consistent under both the random split and time split, albeit smaller as compared with retrieving from the full corpus. 
However, we note that under the time split, if the model is only allowed to retrieve from the historical corpus (the patents for the training data, referred to as ``TS corpus''), \textreact can hardly outperform the baseline. The reason behind this could be that the reactions in new patents are sufficiently different from those in historical patents, placing a predictive barrier for such out-of-distribution generalization. 

\begin{figure}[t]
    \centering
    \includegraphics[width=\linewidth]{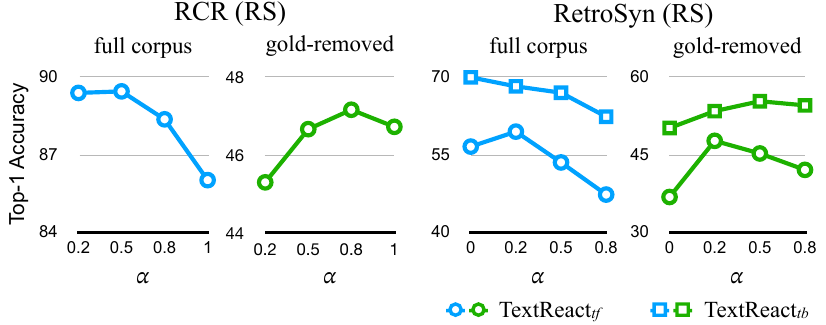}
    \caption{Performance of \textreact with respect to the random sampling ratio $\alpha$ during training. }
    \label{fig:param_alpha}
\end{figure}

\textreact's robust generalization performance is enabled by the sampling strategy we adopted during training. \Cref{fig:param_alpha} illustrates the impact of the random sampling ratio $\alpha$. During training, the predictor is given the gold text paragraph with probability $1-\alpha$, and randomly sampled paragraphs with a probability $\alpha$. The results reveal that the choice of  $\alpha$ plays an important role in the model,  both when retrieving from the full corpus and the gold-removed corpus. In the RCR (RS) dataset, the full corpus evaluation prefers a smaller $\alpha$, while the gold-removed evaluation prefers a larger $\alpha$. Since $\alpha$ controls the probability of using gold input in training, a larger $\alpha$ aligns better with the gold-removed setting. We set $\alpha=0.8$ for the RCR experiments for a good balance between the two settings. In RetroSyn~(RS), however, the model favors a much smaller $\alpha$. We hypothesize that this difference is due to the varying helpfulness of the retrieved texts in different tasks. The texts appear to be more helpful for condition recommendation than retrosynthesis, perhaps due to the existence of common reaction conditions that can be reused across many different reactions of the same type.

\Cref{tab:textreact_ablation} presents our ablation study. \rev{First, \textreact performs significantly better than the model that uses only the input SMILES or retrieved text, illustrating that TextReact effectively integrates both chemistry and text inputs to generate its predictions. Second,} we observe a significant increase in accuracy for \textreact compared to the model trained without the MLM loss. This demonstrates the effectiveness of the auxiliary MLM objective in enhancing the model's learning of the correspondence between chemistry and text inputs. In addition, \textreact benefits from the SciBERT checkpoint, which has been pretrained on scientific text. While \textreact concatenates the top-$k$ neighbors together to make predictions, we compare with a variant that separates the neighbors and ensembles their predictions, similar to RAG \cite{RAG}. This variant yields worse performance (the last row of \Cref{tab:textreact_ablation}), suggesting the benefits of jointly encoding the neighbors.

% We also investigated the effect of the amount of training data on model performance; see \Cref{appendix:analysis} for the discussion.

Additional analyses in \Cref{appendix:analysis} reveal several key findings: (1) \textreact achieves superior performance on the RCR task using only 10\% of the training data compared to the Transformer baseline; (2) \textreact performs better when the retrieved texts describe reactions that bear closer similarity to the input reaction; and (3) \textreact benefits from jointly modeling the input reaction and retrieved neighboring texts.

\begin{table}[t]
    \centering
    \setlength{\tabcolsep}{5pt}
    \resizebox{0.99\linewidth}{!}{%
    \begin{tabular}{lccccc}
    \toprule
         & \multicolumn{2}{c}{RCR (RS)} &&  \multicolumn{2}{c}{RetroSyn (RS)} \\\cmidrule{2-3} \cmidrule{5-6}
         & full & g.r. && full & g.r. \\\midrule
    \textreact          & 88.4 & 47.2 && 59.5 & 47.7 \\
    $\ \cdot$ SMILES-only & 30.0 & 30.0 && 45.9 & 45.9 \\
    $\ \cdot$ text-only & 81.7 & 38.3 && 23.0 & 9.0 \\
    $\ \cdot$ no MLM    & 87.7 & 47.0 && 51.0 & 43.5 \\
    $\ \cdot$ no pretrain & 83.5 & 43.8 && 48.1 & 40.3 \\
    $\ \cdot$ sep. neighbors & 80.7 & 45.2 && 49.4 & 45.3 \\
    \bottomrule
    \end{tabular}
    }
    \caption{Ablation study of \textreact. \rev{We evaluate \textreacttf for the RetroSyn task. } (full: full corpus; g.r.: gold-removed)}
    \label{tab:textreact_ablation}
\end{table}

\section{Conclusion}
This paper presents \textreact, a novel method that augments predictive chemistry with text retrieval. 
We employ information retrieval techniques to identify relevant text descriptions for a given chemistry input from an unlabeled corpus, and supply the retrieved text as additional evidence for chemistry prediction. In two chemistry tasks, \textreact demonstrates strong performance when retrieving from the full corpus, and maintains a significant improvement when retrieving from a harder corpus that excludes the gold texts.

\rev{
Our results highlight the promising potential of incorporating text retrieval methods in the field of chemistry. As chemically similar reactions have similar conditions and outcomes, effectively retrieving and grounding textual knowledge from patents and publications into the chemistry space can significantly enhance the predictive power of computational models.
}

\section{Limitations}
We acknowledge two limitations of this paper. 
First, our experiments focused on two representative chemistry tasks, but we believe that the proposed method can be applied to other tasks and domains that would benefit from the knowledge in the literature. 
Second, we employed a simplified implementation of the retriever and predictor models to demonstrate the effectiveness of retrieval augmentation in chemistry. There is significant room for further improvements, such as using more advanced pretrained models and exploring joint training of the models.

\section{Acknowledgements}
\rev{
The authors thank Wenhao Gao, Xiaoqi Sun, Luyu Gao, and Vincent Fan for helpful discussion and feedback. This work was supported by the NSF Expeditions grant (award 1918839: Collaborative Research: Understanding the World Through Code), the Machine Learning for Pharmaceutical Discovery and Synthesis (MLPDS) consortium, the Abdul Latif Jameel Clinic for Machine Learning in Health, the DTRA Discovery of Medical Countermeasures Against New and Emerging (DOMANE) threats program, the DARPA Accelerated Molecular Discovery program, the NSF AI Institute CCF-2112665, the NSF Award 2134795, the GIST-MIT Research Collaboration grant, MIT-DSTA Singapore collaboration, and MIT-IBM Watson AI Lab.
}

\bibliography{reference}
\bibliographystyle{acl_natbib}

% \newpage
\appendix
\section{Background}
\subsection{Reaction Condition Recommendation}
Reaction condition recommendation is the task of suggesting reaction conditions, such as catalysts, solvents, and reagents, for a chemical reaction. \citet{acscentsci.8b00357} proposed a machine learning model for this task, where the reaction is represented as the difference between the product and the reactant fingerprint vectors, indicating the change of substructures during the reaction, and the reaction conditions are sequentially predicted by the model. Later research studied different methods to represent the reaction, such as MACCS key fingerprints \cite{acs.jcim.9b00313} and graph neural networks \cite{ryou2020graph}, and alternative machine learning formulations, such as multilabel classification \cite{acs.jcim.0c01234} and variational inference \cite{acs.jcim.2c01085}. In this work, we follow \citeauthor{acscentsci.8b00357}'s formulation but use the more advanced Transformer architecture.

\subsection{One-Step Retrosynthesis}
\label{appendix:retro}
Another important task in predictive chemistry is one-step retrosynthesis, which aims to propose reaction precursors (reactants) for target molecules (products). Prior methods can be broadly classified into two categories: \textit{template-based} and \textit{template-free}. Template-based approaches use classification models to predict the reaction template that can be applied to the target \cite{chem.201605499,localretro}, and subsequently employ cheminformatics software to derive the precursors based on the template. Template-free approaches have a more ambitious goal: predicting the precursors directly without relying on a fixed set of reaction templates. Such approaches are usually implemented as graph-to-graph or sequence-to-sequence generation models \cite{acscentsci.7b00303,DaiLCDS19,ShiXG0T20,C9SC03666K,SomnathBCKB21}, and achieve comparable performance to template-based approaches on common benchmarks.
\rev{In this work, we apply \textreact to both template-based and template-free approaches.
}

\begin{figure}
    \centering
    \includegraphics[width=\linewidth]{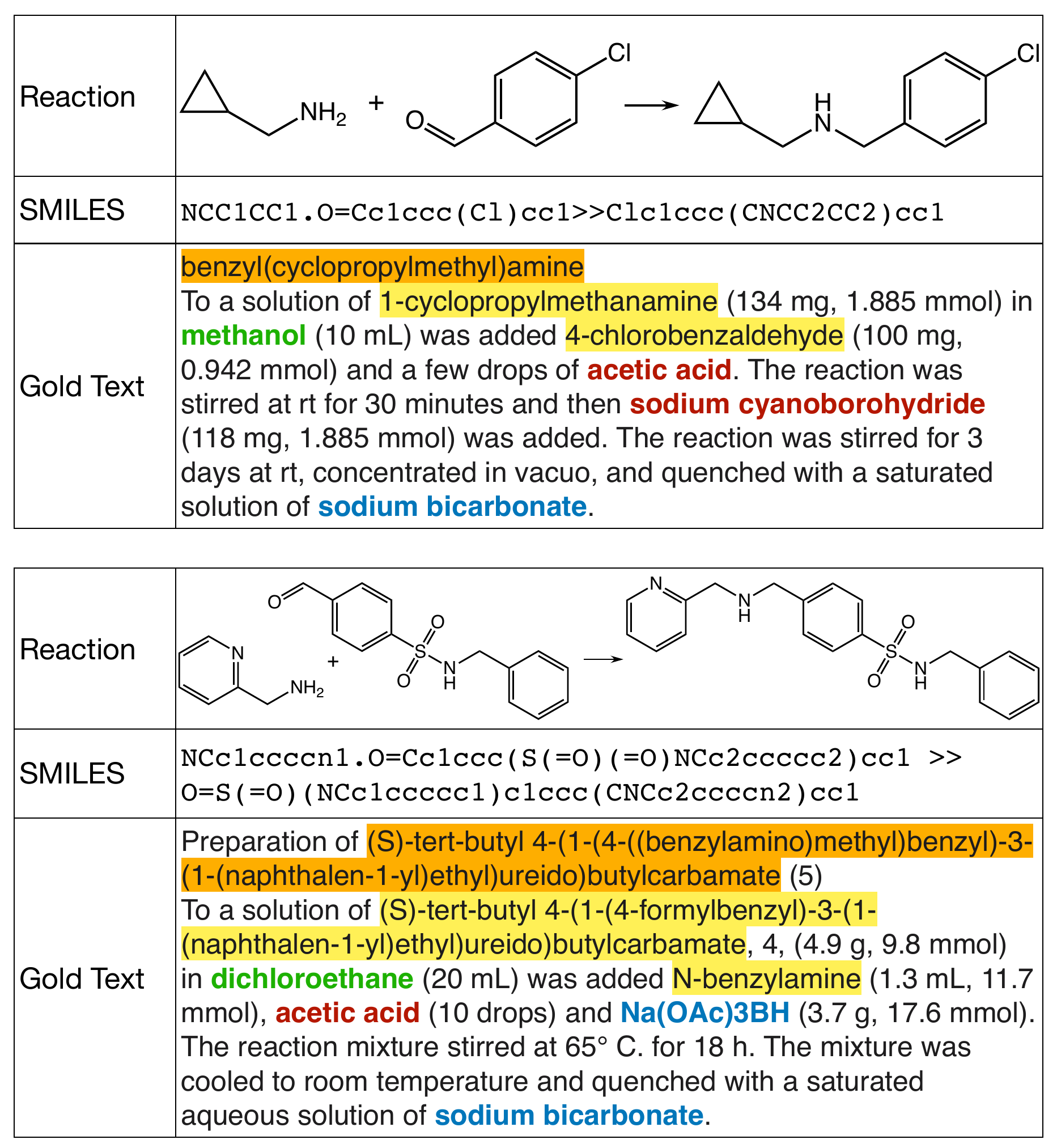}
    \caption{Example reactions and their corresponding gold texts in the USPTO data. Reactants and products are marked in \colorbox{yellow}{yellow} and \colorbox{orange!65}{orange}, while the catalyst, solvent, and reagent are displayed in \textcolor{Maroon}{red}, \textcolor{Green}{green}, and \textcolor{RoyalBlue}{blue}, respectively. }
    \label{fig:goldtext}
\end{figure}

\section{Datasets}
\label{appendix:dataset}

\Cref{fig:goldtext} shows two example reactions and their corresponding gold texts in the USPTO corpus.

\paragraph{Reaction Condition Recommendation}
The dataset is constructed from the public USPTO data \cite{lowe2012extraction}.\footnote{\url{https://doi.org/10.6084/m9.figshare.5104873.v1}} We create two splits:
\begin{itemize}
    \item RCR (RS), where the reactions are \textit{randomly split} into training/validation/testing with a ratio of 80\%/10\%/10\%.
    \item RCR (TS), where we perform a \textit{time split} of the dataset. The reactions collected from patents before 2015 are categorized as the training set, reactions from 2015 as validation, and reactions from 2016 as testing.
\end{itemize}

\paragraph{One-Step Retrosynthesis}
We adopt the USPTO 50K dataset \cite{acscentsci.7b00355}\footnote{\url{https://github.com/coleygroup/openretro/tree/main/data/USPTO_50k}}, which is commonly used in previous works, and create two splits:
\begin{itemize}
    \item RetroSyn (RS), i.e. \textit{random split}, which is the original split of the dataset. 
    \item RetroSyn (TS), i.e. \textit{time split}, where we merge the original training, validation, and testing sets and re-split based on the patent year. Data before 2012 are used for training, data from 2012 and 2013 are used for validation, and data from 2014 and 2015 are used for testing.
\end{itemize}

Our datasets will be publicly available to foster future research in this direction.

\section{Implementation Details}
\label{appendix:implement}

Our experiments are implemented with the Hugging Face Transformers \cite{WolfDSCDMCRLFDS20}, PyTorch Lightning\footnote{\url{https://www.pytorchlightning.ai/index.html}}, Tevatron \cite{tevatron}, and Faiss \cite{johnson2019billion} libraries.

\paragraph{Reaction Condition Recommendation}
The SMILES-to-text retriever consists of a chemistry encoder and a text encoder. We train the retriever by contrastive learning, where each batch contains 512 queries (chemistry inputs), and each query is associated with one positive paragraph and one random sampled negative paragraph from the corpus. Both query and paragraph have a maximum length of 256. We train the retriever for 50 epochs using a learning rate of $\mathrm{1e-4}$ (with 10\% warmup and linear decay). 

The predictor is trained for 20 epochs using a batch size of 128 and a learning rate of $\mathrm{1e-4}$ (with 2\% warmup and cosine decay). The weight of the MLM loss is $\lambda_1=0.1$. We apply data augmentation by generating the SMILES strings with a random order during training. By default, we append the chemistry input with $k=3$ neighboring text paragraphs, and a maximum length of the encoder input is 512. The input SMILES string and text are tokenized with the same SciBERT tokenizer.  For the experiments with $k=1,2,4,5$, we set the maximum length to 256, 384, 768, 1024, respectively.

\paragraph{One-Step Retrosynthesis}
The implementation and hyperparameters for the one-step retrosynthesis experiments largely resemble those of the reaction condition recommendation experiments, with a few differences. The retriever is trained for 400 epochs due to the smaller size of the RetroSyn (i.e., USPTO 50K) dataset. Each query (the SMILES string of the product molecule) has a maximum length of 128. 

\rev{For the predictor, the template-free \textreacttf and template-based \textreacttb architectures are slightly different. While both employ a SciBERT encoder, \textreacttf uses a 6-layer Transformer decoder to generate the output SMILES string, and \textreacttb predicts the reaction template using linear heads on atom and bond representations.}
In both scenarios, the predictor is trained for 200 epochs. The random sampling ratio in training is also set differently ($\alpha=0.2$) as mentioned in \Cref{sec:implementation}.

\section{Further Analysis}
\label{appendix:analysis}

\begin{figure}[tb]
    \centering
    \includegraphics[width=\linewidth]{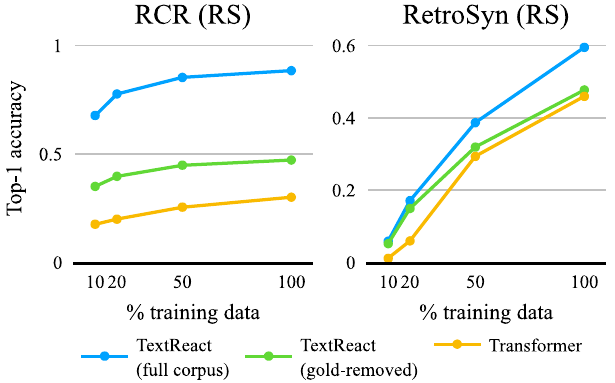}
    \caption{Top-1 accuracies of \textreact and the Transformer baseline vs.\ the
    amount of training data used, as a percentage of the total available training data.
    For RCR, \textreact exhibits strong performance even in low-resource scenarios. \rev{For RetroSyn, we evaluate \textreacttf in this figure.}
    }
    \label{fig:trainnum}
\end{figure}

To investigate the effect that the amount of training data has on
model performance, we trained \textreact in both the RCR and RetroSyn on 10\%, 20\%, 50\%, and 100\% of the available training data under the random split. 
The same was done with the Transformer baseline model for comparison.
The models are evaluated on the full test dataset and the top-1
accuracies are plotted in \Cref{fig:trainnum}.
As expected, both \textreact and the Transformer baseline see a noticeable improvement in
performance as the amount of training data is increased.
In the RCR domain, the full-corpus accuracy and gold-removed accuracy of \textreact when
trained on 10\% of the available training data are 67.7\% and 35.0\%, respectively,
both of which are higher than the accuracy of the baseline as trained
on the \textit{full training set} (30.0\%).
This shows that our model is capable of achieving better performance
with only 10\% as much training data for reaction condition recommendation.
% The effect is particular noticeable in the RetroSyn domain.
% This is likely because the total amount of training data for RetroSyn
% (40k) is already less than 10\% of that for RCR (547k) to begin with, so
% decreasing the amount of training data even further causes model
% performance to suffer greatly.
% These results show the importance of sufficient training data (on the order of
% $\sim 10^5$) for \textreact to perform well.

\begin{figure}[tb]
    \centering
    \includegraphics[width=\linewidth]{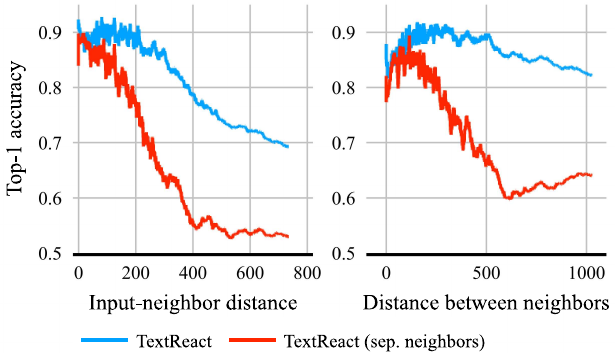}
    \caption{Top-1 accuracies of \textreact and \textreact (sep.\ neighbors)
    on RCR (RS) when retrieving from the full corpus, plotted against
    the average input-neighbor distance and the average distance between neighbors.
    % Accuracies are computed using a sliding window of width 1000 across
    % the test dataset sorted by the variable plotted on the $x$-axis.
    }
    \label{fig:accsim}
\end{figure}

\begin{figure}[t!]
    \centering
    \includegraphics[width=\linewidth]{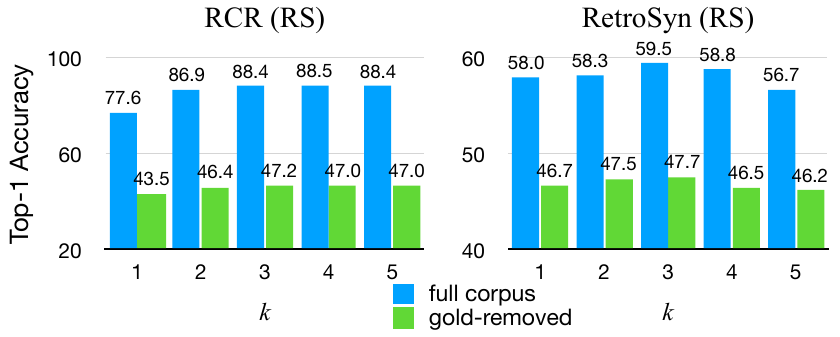}
    \caption{Performance of \textreact with respect to the number of neighbors $k$ retrieved for each input. 
    \rev{We evaluate \textreacttf for RetroSyn. }}
    \label{fig:param_k}
\end{figure}

To study the effect that the retrieved texts have on model performance, we plotted
the top-1 accuracy of \textreact and \textreact (sep.\ neighbors) with respect to
the \textit{average input-neighbor distance}, defined as the average $L_2$ distance between
the fingerprint of the input reaction and the fingerprints of the reactions
corresponding to the retrieved texts (left panel of \Cref{fig:accsim}).
This distance measures the difference between the input reaction
and the reactions described by the retrieved texts. 
The plot shows that model predictions are more accurate when the retrieved
texts correspond to reactions more similar to the input reaction, thus
confirming that \textreact effectively incorporates
information from the retrieved texts when making predictions.

We also plotted the accuracy with respect to the \textit{average distance between neighbors},
defined as the average $L_2$ distance between the fingerprints of the reactions
corresponding to each pair of retrieved texts (right panel of \Cref{fig:accsim}).
This measures how different the retrieved neighbors are from each other.
The graph shows that \textreact's performance does not depend
much on how similar the retrieved neighbors are to each other, whereas the performance of
\textreact (sep.\ neighbors) drops when the retrieved neighbors are more different
from each other. When separately encoding the neighbors, the model does not know how much it should trust each neighbor. 
On the other hand, \textreact does not suffer from the same performance drop,
likely due to the attentions between the retrieved texts, which allow the model to better integrate the information from the neighbors.
This demonstrates the benefits of concatenating the input reaction and
all the retrieved texts together and feeding them into a single Transformer encoder.

\Cref{fig:param_k} analyzes the number of text paragraphs~$k$ that we retrieve for each example. In the RCR~(RS) dataset, there is a notable improvement from $k=1$ to $k=3$, suggesting the benefits of retrieving more neighbors as additional context. In the RetroSyn (RS) dataset, the trend is similar but the improvement is smaller. However, the performance decreases when $k>3$ for both datasets. Therefore, we set $k=3$ for our main experiments.

\end{document}